\documentclass[conference]{IEEEtran}

\usepackage{graphicx}
\usepackage{amsmath,amssymb} 
\usepackage{footmisc}
\usepackage{color}

\ifCLASSINFOpdf
\else
\fi

\begin{document}
%
\title{Improving Text Proposals for Scene Images with Fully Convolutional Networks}

  \author{\IEEEauthorblockN{Dena Bazazian\IEEEauthorrefmark{1}\IEEEauthorrefmark{2},
  		Ra\'ul G\'omez\IEEEauthorrefmark{1}\IEEEauthorrefmark{3}, Anguelos Nicolaou\IEEEauthorrefmark{1}\IEEEauthorrefmark{2}, Llu\'is G\'omez\IEEEauthorrefmark{1}\IEEEauthorrefmark{2}, \\ Dimosthenis Karatzas\IEEEauthorrefmark{1}\IEEEauthorrefmark{2}, Andrew D. Bagdanov\IEEEauthorrefmark{4}}
        
   \IEEEauthorblockA{
    \centerline{ \IEEEauthorrefmark{1} Universitat Aut\`onoma de Barcelona (UAB), Barcelona, Spain}
    \and
     \centerline{ \IEEEauthorrefmark{2}Computer Vision Centre (CVC), Barcelona, Spain} 
     \and
    \centerline{\IEEEauthorrefmark{3}Eurecat - Catalunya Technology Center, unit of  Multimedia Technologies, Barcelona, Spain}
    \and 
    \centerline{\IEEEauthorrefmark{4}Media Integration and Communication Center (MICC), University of Florence, Florence, Italy}}   
  		       
        Email: \{dbazazian, anguelos, lgomez, dimos\}@cvc.uab.es, \\raul.gomez@ce.eurecat.org, andrew.bagdanov@unifi.it
        }

\maketitle

\begin{abstract}
Text Proposals have emerged as a class-dependent version of object proposals -- efficient approaches to reduce the search space of possible text object locations in an image. Combined with strong word classifiers, text proposals currently yield top state of the art results in end-to-end scene text recognition. In this paper we propose an improvement over the original Text Proposals algorithm of \cite{Gomez16}, combining it with Fully Convolutional Networks to improve the ranking of proposals. Results on the ICDAR RRC and the COCO-text datasets show superior performance over current state-of-the-art.
\end{abstract}


%
\IEEEpeerreviewmaketitle

\section{Introduction}
Text understanding in unconstrained scenarios, such as text in scene images or videos, has attracted a lot of attention from the computer vision community. Though extensively studied in recent years, localizing text in complex images remains quite challenging.

Object proposal techniques have emerged as an efficient approach to reducing the search space of possible object locations in an image by generating candidate class-independent object locations and extents. Such generic object proposal methods are typically designed to detect single-body objects, and are not appropriate for text detection which aims to detect groups of disjoint, atomic objects (characters or text strokes). The Text Proposals technique was recently suggested by Gomez et al.~\cite{Gomez16} as an alternative, class-specific object proposal method that takes into account the particular characteristics of text.  Text Proposals combined with a strong word classifier produces state-of-the-art results in end-to-end scene text recognition~\cite{Gomez16}.

In this paper we look at how the power of Fully Convolutional Networks (FCN) can be leveraged to improve Text Proposals. The robustness of FCN stems from the use of convolutional layers instead of fully-connected layers as in conventional networks, which results in spatial information being preserved in the final discriminative layer of the network. FCNs have recently been applied to text detection with notable results~\cite{Zhang16a}. The output of such a network is a coarse heatmap of locations of interest, which must be further processed to produce a proper localization result.

The Text Proposals algorithm, on the other hand, produces a short list of bounding boxes of probable text locations. The method follows a strategy based on an over-segmentation of the image and the efficient construction of meaningful groupings of regions, resulting in very accurate localization. Yet, like all object proposals approaches, it is optimized for high recall and produces a considerable number of false positives.

In this paper we train an FCN for text detection and use resulting heatmap to rerank the set of possible text locations obtained by the Text Proposals algorithm. Different ranking strategies are explored that yield state-of-the-art text proposals results over several datasets.

The main contribution of this paper is the development and evaluation
of different ranking strategies for combining the Text Proposals approach with the FCN obtained text probability
scores, yielding significantly better recall rates while considering an order of magnitude fewer proposals.


The remainder of the article is organized as follows. Section \ref{Sec:Related Work} presents related work, followed by a description of our approach and architecture in Section \ref{sec:PropsMeth}. Section \ref{Sec:Experiment and Results} reports on the experimental results of our approach, and conclusions are drawn in section \ref{Conclusion}.

\section{Related Work}
\label{Sec:Related Work}
Text detection and recognition in scene images are gaining increasing attention from the computer vision community. In spite of the immense effort which has been devoted to improving the performance of text detection in unconstrained environments, it is still quite challenging due to the diversity of text appearance and geometry in the scene and the highly complicated backgrounds.

Text detection and localization algorithms are divided in two main approaches.
First, connected component-based methods perform image segmentation and then classify and group the detected components into text candidates~\cite{Kang14,Li14,Yao12, Yin15,Yin14}. Following this idea, but in the other way around, is~\cite{Gomez13} where the interested components are detected by employing grouping.
Second, sliding-window methods perform region classification at multiple scales and follow standard post-processing to derive precise text locations~\cite{Jaderberg14,Tian15,Wang11,Zhang15a,Wang12}.
Most top-ranking approaches \cite{Koo13,Yin15,Yin14,Wang12,Neumann13}
in the latest editions of the ICDAR Robust Reading Competition~\cite{Karatzas15} are connected component-based methods based on variants of the Maximally Stable Extremal Regions (MSERs) algorithm~\cite{Matas04}.
The MSER algorithm is particularly suited to text detection as it efficiently leverages local contrast patterns. Different approaches in this category include the Stroke Width Transform (SWT)~\cite{Epshtein10} and recent variants like the Stroke Feature Transform (SFT)~\cite{Huang13}. Connected component-based approaches permit early fusion with the language model at the time of filtering components and forming words. A recent example~\cite{Mishra16} proposes a model which combines bottom-up cues from individual character detection and top-down cues from a lexicon.

CNN classifiers are increasingly incorporated in text detection approaches. Wang et al.~\cite{Wang12} employed CNN models for text/non-text classification. 
He et al.~\cite{He15} and Huang et al.~\cite{Huang14} exploit a CNN model to filter out non-character MSER components, while CNN based region classification has been repeatedly employed for sliding-window approaches~\cite{Jaderberg14,Wang12}.
Tian et al. proposed TextFlow~\cite{Tian15} for character classification, which simplifies multiple post-processing steps by utilizing a minimum cost flow network~\cite{Wang12}. These approaches all designed features for detecting character candidates, and they apply a pre-trained CNN model to assign a score to each candidate. On a different line, Zhang et al.~\cite{Zhang15a} trained two classifiers that work at character level and text region level, respectively.

Despite the immense success of CNN models for tasks such as character classification or word-spotting once text regions are localized, the problem of text localization still poses significant challenges.
To this end, the use of generic object proposal techniques for scene text understanding has been exploited by Jaderberg et al.~\cite{Jaderberg16}. Their end-to-end pipeline combines the EdgeBoxes~\cite{Zitnick14} object proposals algorithm and a trained aggregate channel features detector~\cite{Dollar14} with a powerful deep Convolutional Neural Network for holistic word recognition. Their method uses a CNN-based bounding box regression module on top of region proposals in order to improve their quality before word recognition is performed.

The use of generic object proposals is not optimal when text is to be detected, as demonstrated in~\cite{gomez2015object}. Gomez et al.~\cite{Gomez16} propose instead a text-specific object proposals method that is based on generating an hierarchy of word hypotheses based on a similarity region grouping algorithm. Contrary to generic object proposal algorithms, this approach takes into account the specific characteristics of text regions that are fundamentally different from the typical notion of single-body object as normally used in for object detection. Replacing the overly complicated pipeline used for text localization in Jaderberg~\cite{Jaderberg16} with the single step localization offered by~\cite{Gomez16} is currently topping the state of the art for end-to-end methods on the demanding Challenge 4 (incidental text) of the Robust Reading Competition.


On the other hand, Fully Convolutional Networks (FCNs)~\cite{Long15} have recently attracted considerable attention from the robust reading community~\cite{Zhang16a,Zhang16b,He16,Gupta16}. FCN-based methods replace fully-connected layers with convolutional layers which allows them to preserve coarse spatial information which is essential for text localization tasks. Zhang et al.~\cite{Zhang16a} integrated semantic labeling by FCN with MSER to provide a natural solution for handling text at arbitrary orientations. In a parallel work Zhang et al.~\cite{Zhang16b}
designed a character proposal network based on an FCN which simultaneously predicts characterness scores and refines the corresponding locations. The characterness score is used for proposal ranking.

He et al.~\cite{He16} proposed a Cascaded Convolutional Text Network (CCTN) that combines two customized convolutional networks for coarse-to-fine text localization. The CCTN detects text regions roughly from a low-resolution image, and then accurately localizes text lines from each enlarged region. In addition, to precise text detection, Gupta et al.~\cite{Gupta16} presented an extreme variant of Hough voting, inspired by Fully-Convolutional Networks~\cite{Long15} and the YOLO technique~\cite{Redmon15}. The algorithm of YOLO follows the idea of Hough voting, but avoids voting and uses instead individual predictions directly. Therefore, it allows the framework to pick up contextual information which is useful in detecting small text occurrences in natural scene images.

In this paper we take into account the respective advantages of FCN models and Text Proposals in order to propose an improved text proposals algorithm. 
We employ text probability scores obtained through FCN over individual proposals for ranking, efficiently suppressing non-text proposals.
Contrary to Zhang et al.~ \cite{Zhang16a} who propose a coarse-to-fine method to obtain bounding boxes from the FCN heatmap
with the purpose of maximizing localization accuracy, we employ the FCN heatmap to produce text proposals in order to maximize recall by producing the necessary number of proposals in an efficient way.
We choose to use the FCN result to re-rank text proposals obtained by~\cite{Gomez16}, and thereby improving the performance of the text proposals algorithm.


\section{The Proposed Method}
\label{sec:PropsMeth}
\begin{figure}
\centering
\includegraphics[height=1.73cm]{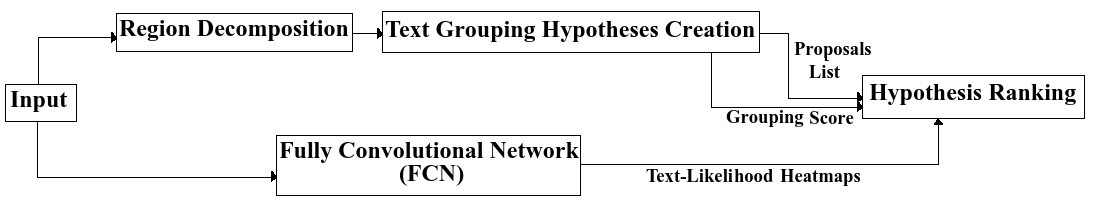}
\caption{Text proposals produce a list of hypotheses for text locations as well as their associated grouping score, while a text detection FCN is simultaneously applied to the image to obtain per pixel text-probability. Text hypotheses are then re-ranked based on the FCN information to obtain the final hypotheses ranking.}
\label{fig:Pipeline}
\end{figure}

In this section we introduce our text proposal framework. The pipeline, shown in Figure~\ref{fig:Pipeline}, comprises three steps. First, a \textbf{text proposal stage} decomposes an image into regions and creates the text group hypotheses based on~\cite{Gomez16} (see Section~\ref{subsec:Text-Proposal}). In parallel, a \textbf{pixel-wise prediction by FCN} stage estimates the pixel-level text probability for the whole image (see Section \ref{subsec:FCN}). Finally, \textbf{Hypotheses Ranking} leverages the FCN predictions to re-rank the text hypotheses (see Section \ref{subsec:Combining TxtProp and FCN}). The remainder of this section discusses these three components in detail.

\subsection{Text Proposals}
\label{subsec:Text-Proposal}
In our framework we use the Text Proposals algorithm of Gomez et al.~\cite{Gomez16}, which is divided in three main steps: (1) an initial over-segmentation of the input image from which we obtain a set of connected components; (2) the creation of text
hypotheses through several bottom-up agglomeration processes; and (3) a ranking strategy that prioritizes the best text proposals. 

Region decomposition, based on the Maximally Stable Extremal Regions (MSER) algorithm, aims to detect the atomic parts that will give rise to subsequent text groupings. The subsequent grouping process builds an hierarchy of groupings of the initial set of MSER regions according to similarity cues such as intensity and color mean and stroke width.

The groupings are first classified as text/non-text based on features
efficiently computed in an incremental way along the hierarchy
structure. Once non-text groupings have been filtered out, remaining
groupings are scored based on a text structure quality score that
stems from perceptual organization principles and the notion of
perceptual meaningfulness of the particular grouping, and form the
resulting text proposals. For further information on the specifics of
the Text Proposal algorithm, please see~\cite{Gomez16}.

\begin{figure}
\centering
\includegraphics[height=1.60cm]{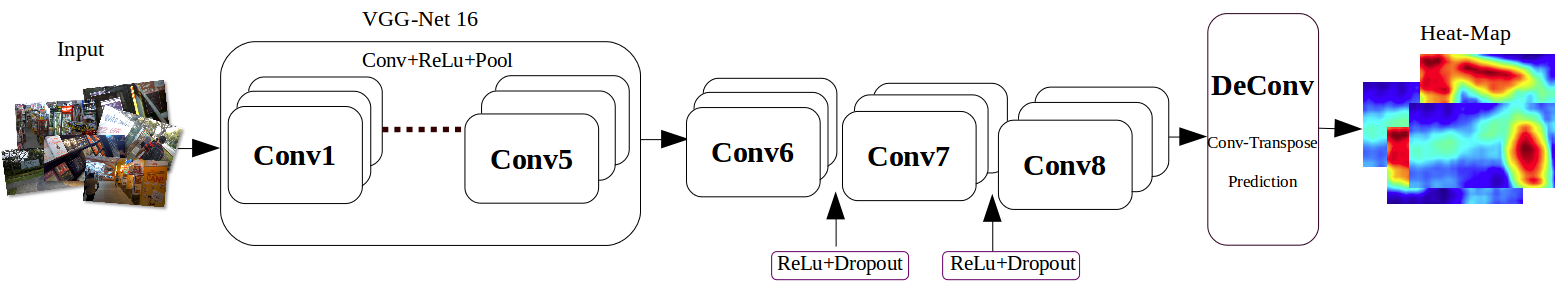}
\caption{\scriptsize Architecture of the Fully Convolutional Network (FCN) used in our pipeline based on~\cite{Long15}.}
\label{fig:FCN}
\end{figure}

\subsection{FCN-based Text Prediction}
\label{subsec:FCN}
Fully convolutional networks (FCNs) are designed to provide pixel-level predictions~\cite{Long15}. Every layer in an FCN computes a local operation on relative spatial coordinates. Since there is no fully-connected layer, it is possible to use FCN on variable dimension images and produce an output of the corresponding input dimension as well as preserving coarse spatial information of the image which is essential for the text detection tasks. 
Hence, in this research we use an FCN to perform per-pixel prediction and estimate a \emph{text heatmap} for the input image.
The architecture of our network is based on~\cite{Long15} and is shown in Figure~\ref{fig:FCN}.

\begin{figure}
\centering
\includegraphics[height=1.62cm]{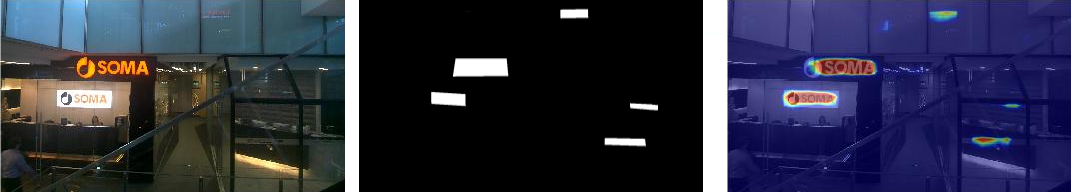}
\caption{An example image from our training set, the corresponding ground truth text mask, and an FCN heatmap superimposed on the original image. The heatmap indicates the degree of ``textness'' estimated by the FCN.}
\label{fig:InMaskHeatMap}
\end{figure}

First, we transform the pre-trained VGG network~\cite{Simonyan15} into a fully convolutional form following Long et al.~\cite{Long15}. The original FCN network was designed for the semantic segmentation of images into the twenty classes of the PASCAL VOC dataset. Thus, we customized the network to our purpose of performing text/none-text segmentation. We apply softmax normalization on the FCN output in order to employ it as text probability for the subsequent hypotheses ranking step. In Figure~\ref{fig:InMaskHeatMap} we show an example image from our training dataset, along with the ground truth text annotations and text heatmap output by our FCN.

\subsection{Hypotheses Ranking}
\label{subsec:Combining TxtProp and FCN}

Once text proposals and text heatmaps are computed then we rank the set of proposals according to three different ranking strategies as explained next.


The \textbf{baseline} (BAS) strategy ranks regions using the grouping quality score provided by the Text Proposals algorithm~\cite{Gomez16} as explained in section~\ref{subsec:Text-Proposal} (does not re-rank text proposals using the FCN heatmap).

The \textbf{mean text probability} (MTP) strategy ranks regions using the mean text probability, obtained from the FCN heatmap averaging over the proposed text region. FCN heatmaps provide only coarse information about text localization. Directly using this information to rank bounding boxes can be tricky as smaller regions corresponding to the inner parts of a text block might appear to be better candidates than the full text block bounding box that covers the extents of the text including lower probability pixels at the borders. Indeed, using the average FCN score over text hypotheses as a ranking mechanism has the adverse effect that small patches within text regions are ranked better than full regions corresponding to words or lines.
On the other hand, the grouping quality score 
prioritizes regions with high probability to be words or text lines, but it produces many false positives.

Therefore, we introduce a \textbf{suppression} (SUP) strategy, 
which optimally combines these two complementary behaviors.
The idea is to suppress the text proposals which have low mean text probability, in order to filter false positives. 
Subsequently, the rank of the remaining regions is established according to the grouping quality score which prioritizes well-structured text blocks. 
The smaller regions corresponding to the inner parts of text blocks are not suppressed by their FCN probability, however, they are generally ranked low based on their grouping quality score as they only cover a small set of atomic parts. 
The suppression strategy allows us to discard a significant number of false positives, resulting in superior detection with a smaller number of proposals while preserving the high recall rate.

\section{Experimental Results}
\label{Sec:Experiment and Results}

In this section we describe experiments we performed to evaluate the proposed method. First, we detail the datasets we used. Afterwards, we quantify the effect of training FCN model over different datasets.
Furthermore, we compare between the baseline and proposed ranking strategies, and finally evaluate the effect of different suppression thresholds.

\subsection{Datasets and Experimental Setup}

\subsubsection{Datasets}
\label {Sec:Datasets}
The experiments of our proposed methods were performed on three current text detection datasets covering a range of conditions: ICDAR-Challenge4, SynthText in the Wild, and COCO-Text.

\begin{itemize}
\item \textbf{ICDAR-Challenge4:}
ICDAR Challenge 4 focuses on incidental scene text~\cite{Karatzas15}. Incidental scene text refers to text that appears in the scene without the user having taken any specific prior action to cause its appearance or to improve its positioning/quality in the frame. While focused scene text (explicitly photographed by the user) is the expected input for applications such as translation on demand, incidental scene text represents another wide range of applications linked to wearable cameras or massive urban captures where the acquisition process is difficult or undesirable to control. This challenge has 1000 images with publicly available ground truth annotations. We randomly selected and set aside 200 of these images to use for testing and used the rest for training.

\item \textbf{SynthText in the Wild:}

SynthText in the Wild 
is a synthetic large scale dataset that provides detailed ground-truth annotations and scalable alternatives to annotating images manually~\cite{Gupta16}. It comprises 858,750 images, which are divided into 200 different themes. Each theme consists of about 4000 images of text synthetically generated onto almost 40 different backgrounds. For training we considered 15000 random images of the SynthText Dataset with different backgrounds from various themes. Since this dataset is synthetic and it was not use for the validation by the original authors. Therefore, we just consider it for the training stage and not for testing.

\item \textbf{COCO-Text:}
The COCO-Text dataset~\cite{Veit16} is based on the MS COCO dataset, which contains images of complex everyday scenes. 
The COCO-Text dataset contains non-text images, legible text images and illegible text images. A training set and a validation set are defined, from which we consider only those images containing at least one instance of legible text. Notice that some of that images may also contain illegible text instances. That is a total of 22184 images for training and 7026 images for validation. We evaluate on the validation set, since the test set of COCO-Text is withheld. 

\end{itemize}

\subsubsection{Training Process}
In this work we trained the FCN network for the task of text detection on the three datasets described in section~\ref{Sec:Datasets}.  
The training for each of the train sets for these datasets is done by fine-tuning a VGG16 network pre-trained on Imagenet~\cite{Simonyan15}.

The FCN networks are all trained using Stochastic Gradient Descent with mini-batches of 20 images, a momentum of 0.9, L2 weight-decay of 5\textsuperscript{-4}, and a  fixed learning rate of 10\textsuperscript{-4}. We apply dropout after the Conv6 and Conv7 layers with a rate of 0.5 as shown in Figure~\ref{fig:FCN}.


\subsection{Performance Evaluation}
The time consumed to generate the FCN heatmaps without taking into account reading the image from disk and writing the heatmap to disk 
is 0.15 seconds per image. All experiments were conducted on a system with a GPU GeForce GTX TITAN and an Intel\textsuperscript{\textregistered} Core\textsuperscript{\tiny TM}2 Quad CPU Q9300@ 2.50GHz 
processor.

The quantitative results are shown in Figure ~\ref{fig:Scoring-strategies} and Table~\ref{tab:ComparisonBetweenStrategies}. Each method is coded as [Ranking Strategy]\_[FCN Training Set] while for the Suppression Strategy, we also indicate the suppression threshold used.

\subsubsection{Effect of Training on different datasets}
We trained 
the FCN using all three different training sets, and we evaluate each trained model on both ICDAR-Challenge4 and COCO-Text. We report results using the mean text probability (MTP) ranking strategy (MTP\_COCO, MTP\_ICDAR and MTP\_SYNTH). Examining the corresponding plots it can be seen that adapting on different training sets has minor influence on the ranking results and the quality of heatmaps produced while there is no observed difference between training on real (COCO, ICDAR) and synthetic (SYNTH) data.

At the same time, it can be observed that all curves based on the mean text probability ranking strategy (using only the FCN heatmap information to rank the regions returned by the Text Proposals method) perform notably worse than the baseline.

\begin{figure}
\begin{center}
\begin{tabular}{ccc}
\includegraphics[width=0.140\textwidth]{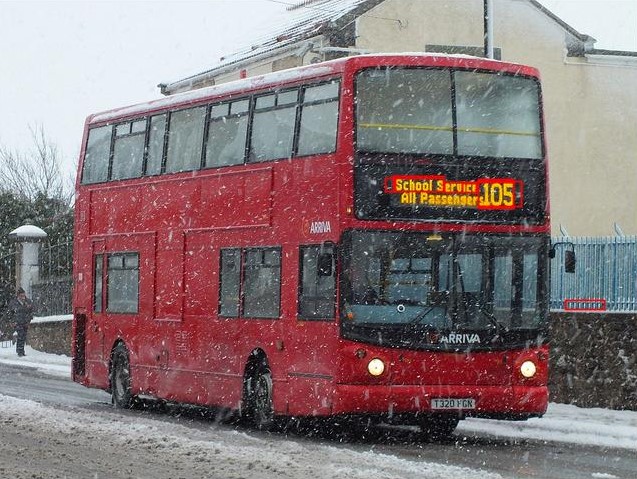} &
\includegraphics[width=0.140\textwidth]{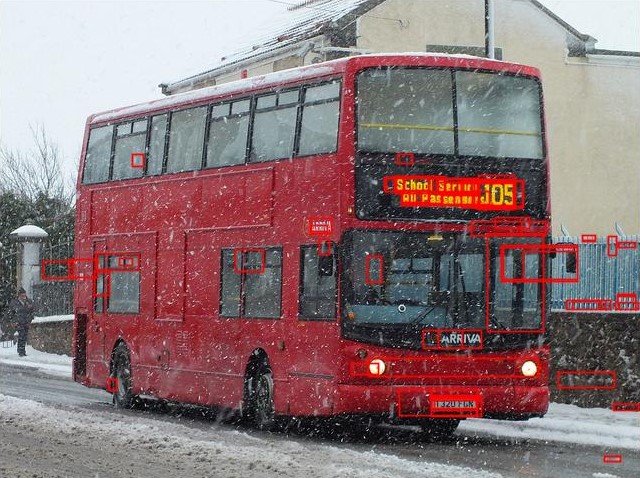} &
\includegraphics[width=0.140\textwidth]{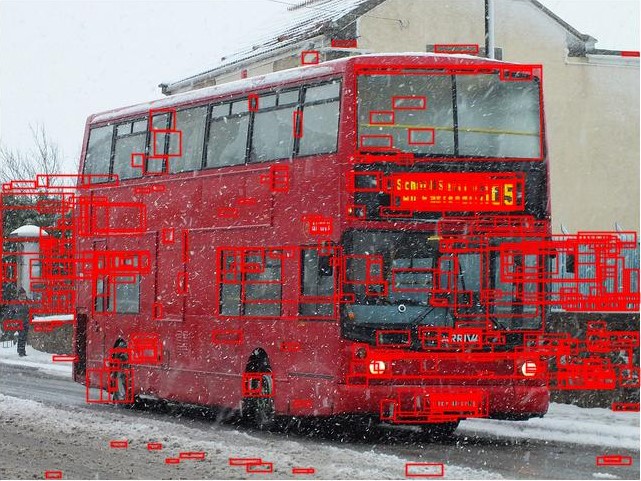} \\
\includegraphics[width=0.140\textwidth]{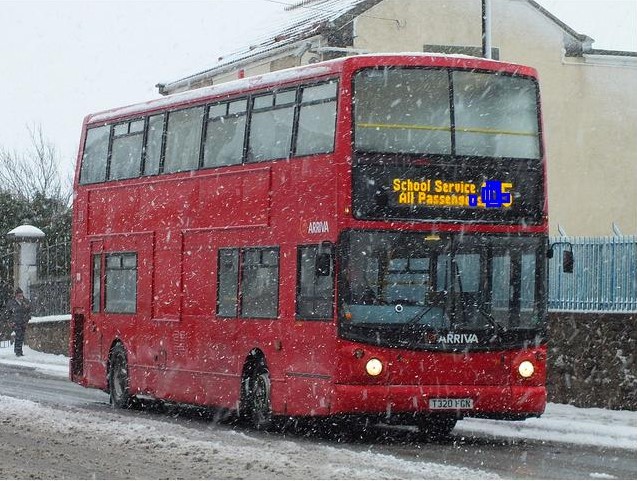} &
\includegraphics[width=0.140\textwidth]{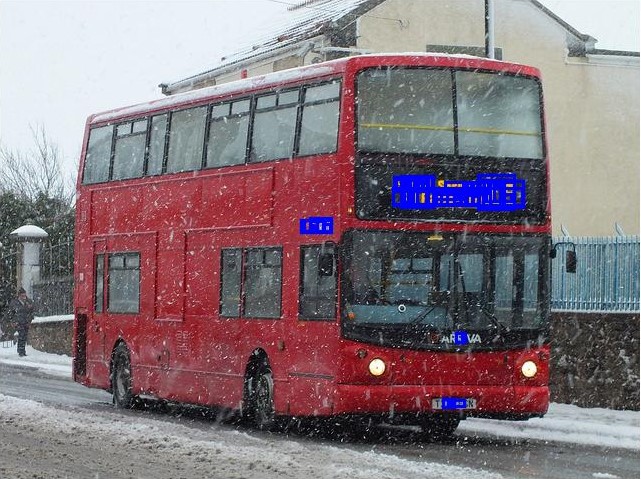}&
\includegraphics[width=0.140\textwidth]{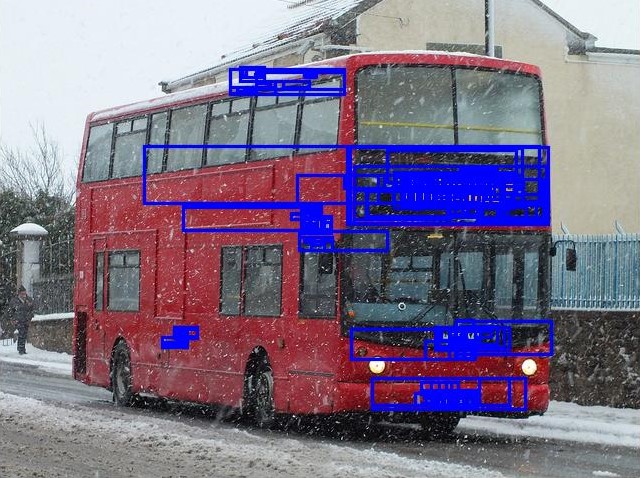} \\
\scriptsize(a)Top 10 proposals & \scriptsize(b)Top 100 proposals & \scriptsize(c)Top 500 proposals
\end{tabular}
\end{center}
\caption{Examples of top proposals by strategy. Top: red regions are those ranked high by Baseline strategy (BAS). Bottom: blue boxes are those ranked high by the FCN.}
\label{fig:topRank-TP-FCN}
\end{figure}

Looking at the top-N proposals for both the Baseline(BAS) and the MTP strategies in Figure~\ref{fig:topRank-TP-FCN} it can be appreciated that by using solely the FCN information, the method selects more relevant regions, however promotes small parts of text blocks. While the Baseline(BAS), making use of the grouping quality score, then returns regions that better match the extent of the text block, but also produces a lot of false positives.

\subsubsection{Suppression Strategy}
\label{Suppression-threshold}
The suppression strategy has a superior performance over the FCN information by combining it with the grouping quality score.

As the training set produces little difference in the quality of the model, here we report results using the FCN model trained on the COCO-Text dataset only and focus on the selection of the suppression threshold (using models trained on ICDAR or SYNTH produce very similar results).

\begin{figure}
\begin{center}
\includegraphics[width=0.479 \textwidth]{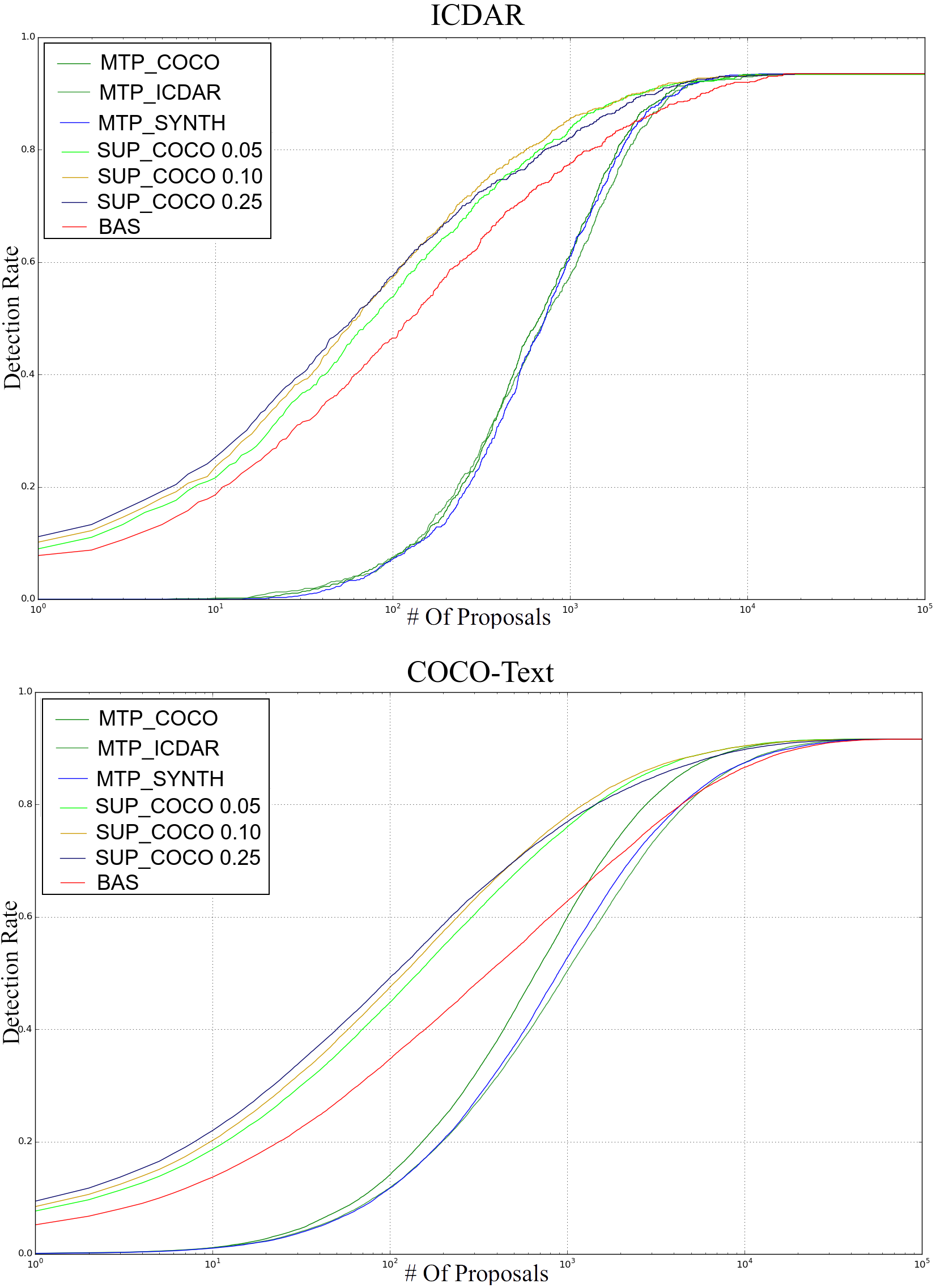}
\end{center}
\caption{Evaluating different detection rate of different strategies on ICDAR-Challenge4 (top) and COCO-Text (bottom) datasets.} 
\label{fig:Scoring-strategies}
\end{figure}

Figure~\ref{fig:Scoring-strategies} shows results employing different thresholds. A threshold of 0.1 yielded the best results in our experiments, since according to Figure~\ref{fig:Scoring-strategies} and Table~\ref{tab:ComparisonBetweenStrategies}, it demonstrates that by considering smaller threshold the detection rate gets less. 

Moreover, in Table~\ref{tab:ComparisonBetweenStrategies}, we present indicative quantitative results for detection rate at 10, 100 and 1000 proposals, which represents a reasonable trade-off for real-life applications.

\begin{table}
\caption{Detection rate of different strategies using 10, 100 and 1000 proposals on ICDAR-Challenge4 and COCO-Text datasets.}
\label{tab:ComparisonBetweenStrategies}
\begin{center}
\begin{tabular}{|c||c|c|c||c|c|c|}
\hline
 & \multicolumn{3}{|c||} {ICDAR} & \multicolumn{3}{|c|} {COCO-Text} \\
 \hline
 \# of Proposals & 10 & 100 & 1000 & 10 & 100 & 1000 \\
\hline\hline
SUP\_COCO 0.10 & 0.23 & \textbf{0.57} & \textbf{0.85} & 0.20 & 0.47 & \textbf{0.78}  \\
\hline
SUP\_COCO 0.05 & 0.21 & 0.53 & 0.83 & 0.18 & 0.44 & 0.76 \\
\hline
SUP\_COCO 0.25 & \textbf{0.25} & \textbf{0.57} & 0.82 & \textbf{0.21} & \textbf{0.49} & 0.77 \\
\hline
BAS & 0.18 & 0.46 & 0.77 & 0.13 & 0.34 & 0.63 \\
\hline
MTP\_COCO & 0.00 & 0.07 & 0.61 & 0.01 & 0.15 & 0.60 \\
\hline
MTP\_SYNTH & 0.00 & 0.07 & 0.61 & 0.01 & 0.11 & 0.52 \\
\hline
MTP\_ICDAR & 0.00 & 0.07 & 0.57 & 0.01 & 0.11 & 0.50 \\
\hline
\end{tabular}
\end{center}
\end{table} 
\subsection{Qualitative Results}
\begin{figure*}
\begin{center}
\begin{tabular}{cccc}
\includegraphics[width=0.265\textwidth]{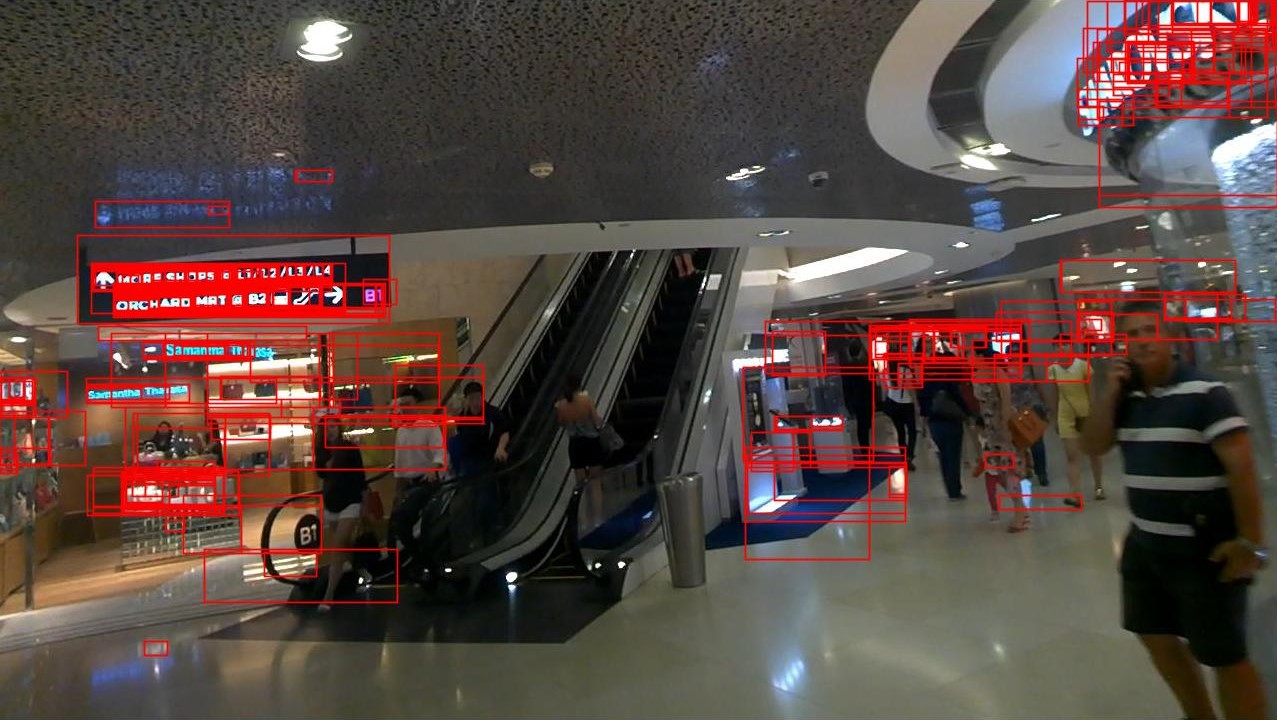}&
\includegraphics[width=0.265\textwidth]{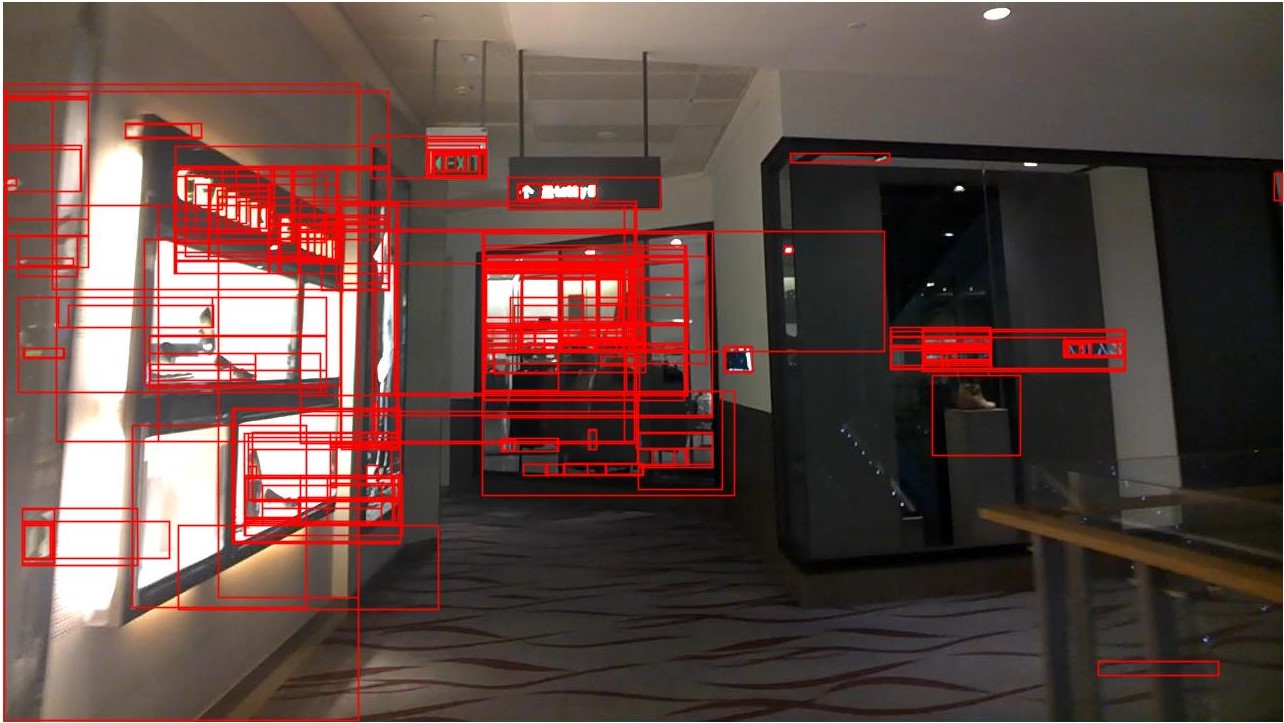} &
\includegraphics[width=0.205\textwidth]{results/coco_5169_tp_500}&
\includegraphics[width=0.116\textwidth]{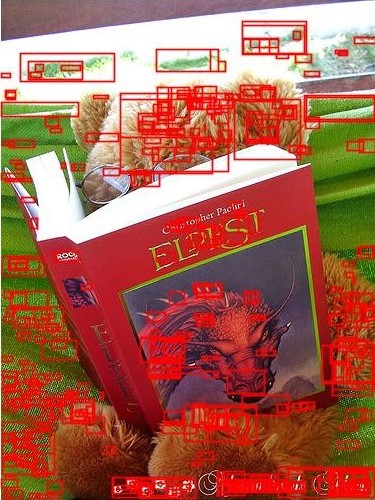}\\

\includegraphics[width=0.265\textwidth]{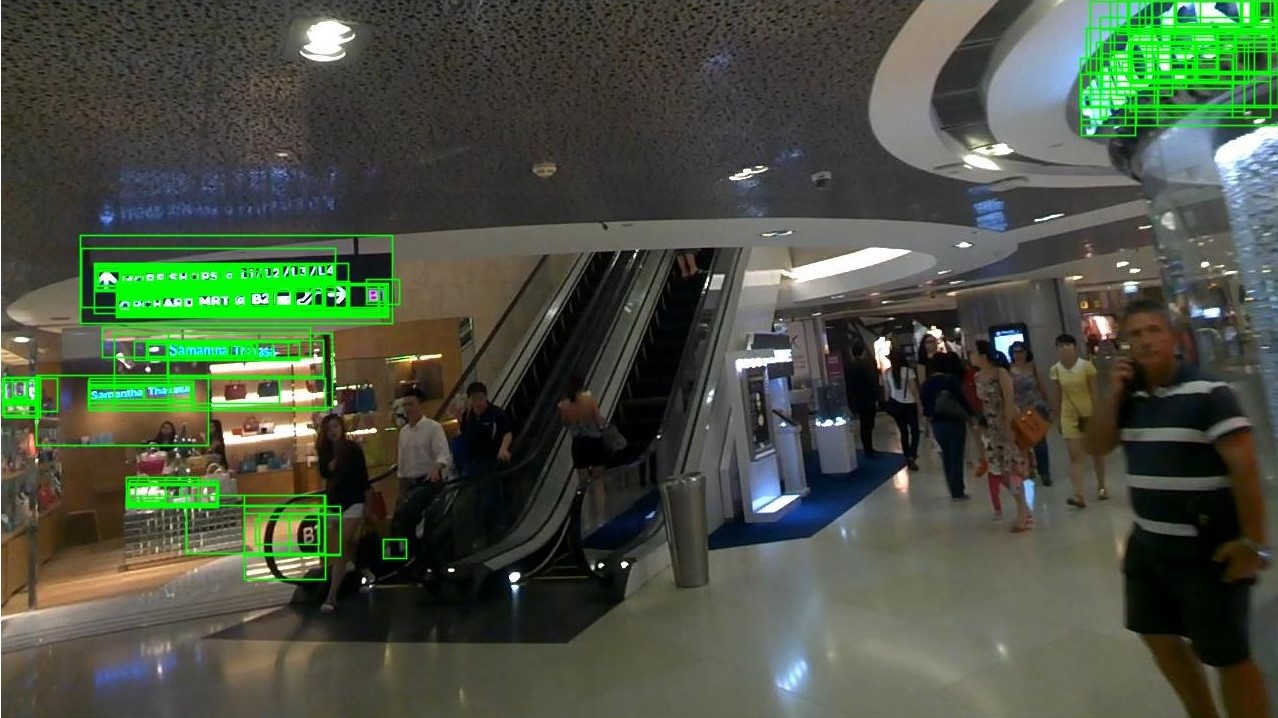} &
\includegraphics[width=0.265\textwidth]{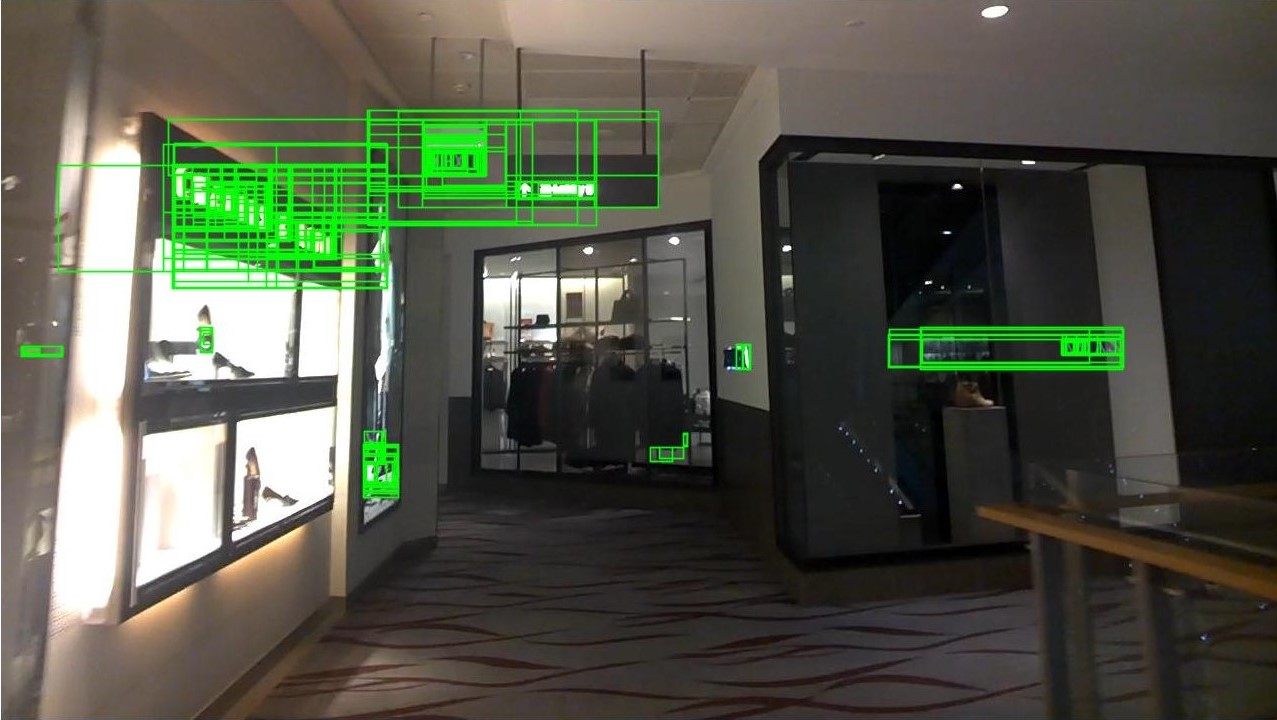} &
\includegraphics[width=0.205\textwidth]{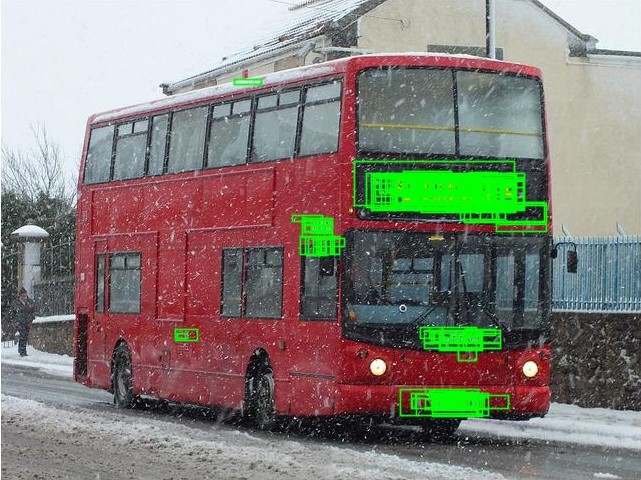}& 
\includegraphics[width=0.116\textwidth]{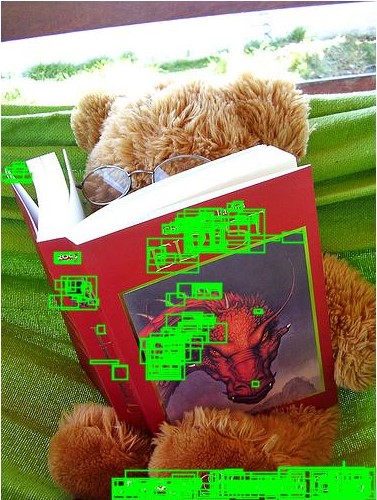}

\end{tabular}
\end{center}
\caption{ICDAR-Challenge4 and COCO-Text ranking examples. In red, regions ranked by Baseline strategy (BAS), and in green those ranked by the suppression strategy (0.10). The top 500 ranked regions by each strategy are drawn.}
\label{fig:topRank-icdar-coco}
\end{figure*}

Figure~\ref{fig:topRank-icdar-coco} illustrates the bounding boxes of the top 500 ranked regions using the baseline and suppression (threshold 0.10) ranking strategies on selected images from the COCO-Text dataset. Similarly, in Figure~\ref{fig:topRank-icdar-coco} we give examples from ICDAR-Challenge4.
The proposed ranking strategy is capable of detecting varied types of text, including different fonts, handwritten text, various orientations or deformations, diverse text length, and also different languages.

\begin{figure*}
 	\centering
 	\begin{minipage}[t!]{.5\textwidth}
\begin{center}
\begin{tabular}{ccc}
\includegraphics[width=0.275\textwidth]{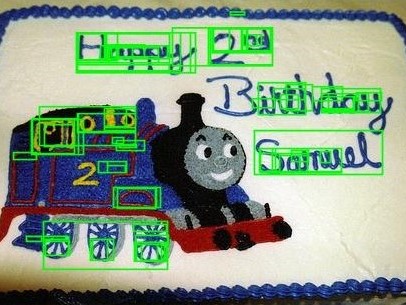} &
\includegraphics[width=0.275\textwidth]{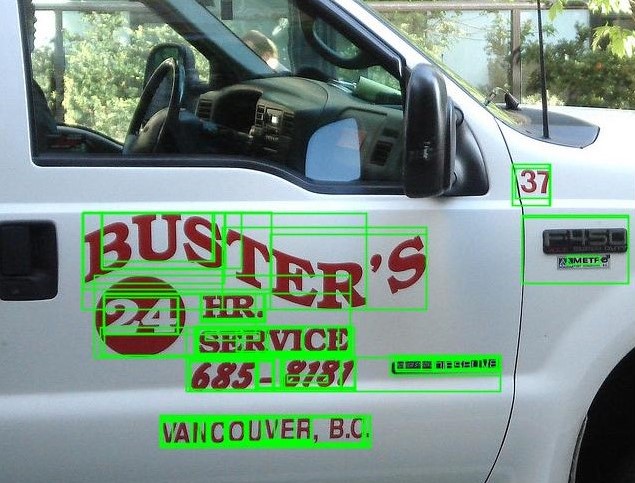} &
\includegraphics[width=0.275\textwidth]{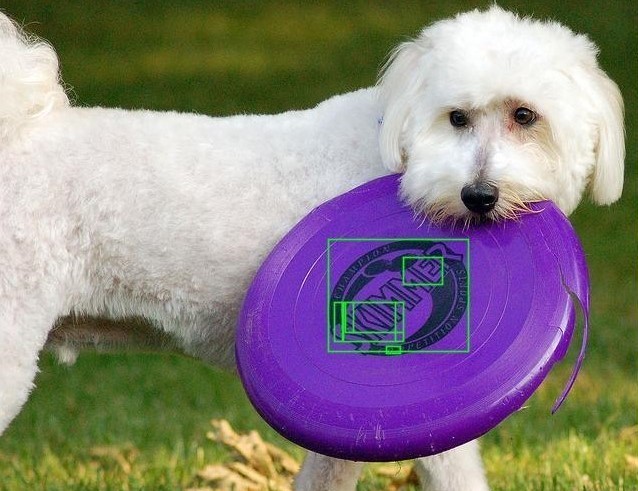} \\
\includegraphics[width=0.275\textwidth]{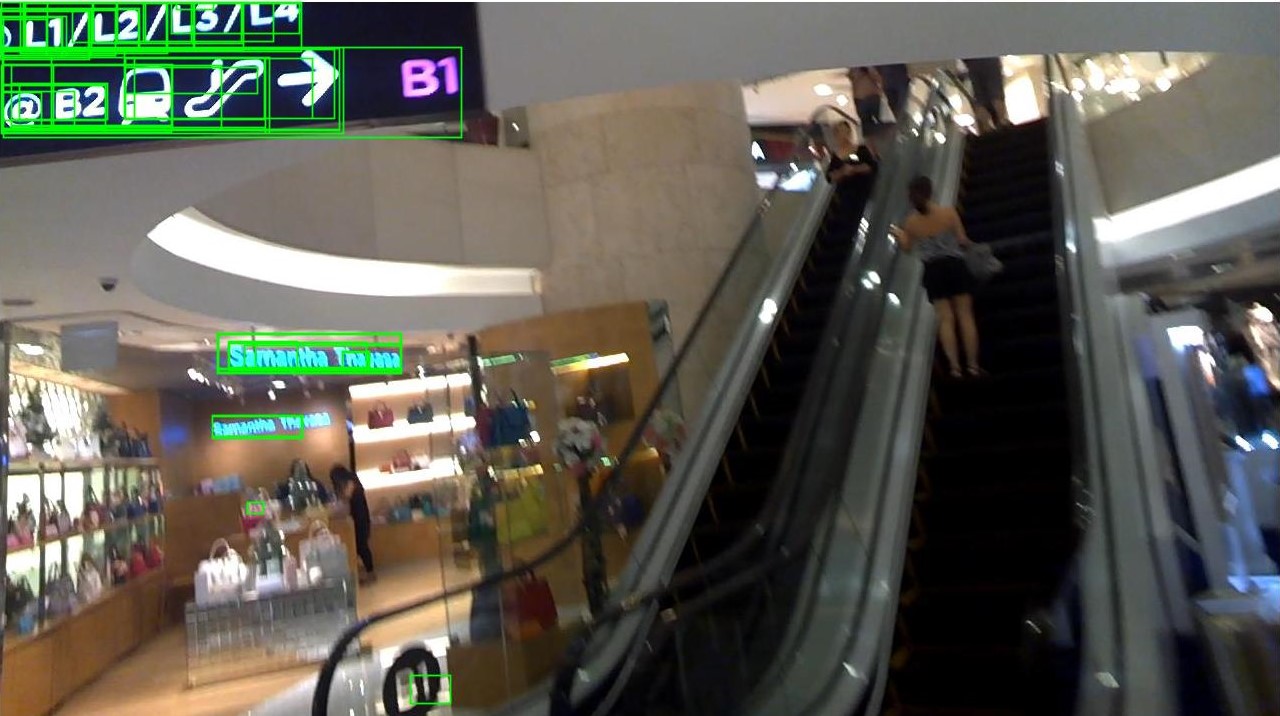} &
\includegraphics[width=0.275\textwidth]{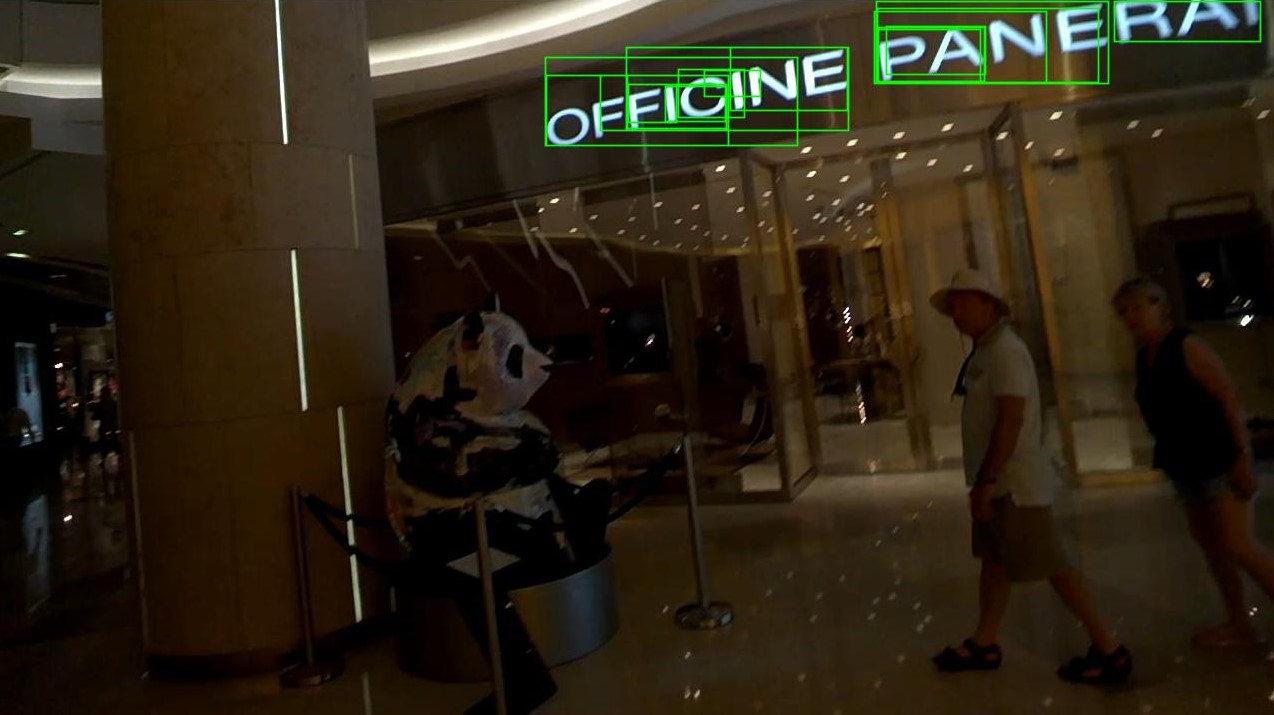} &
\includegraphics[width=0.275\textwidth]{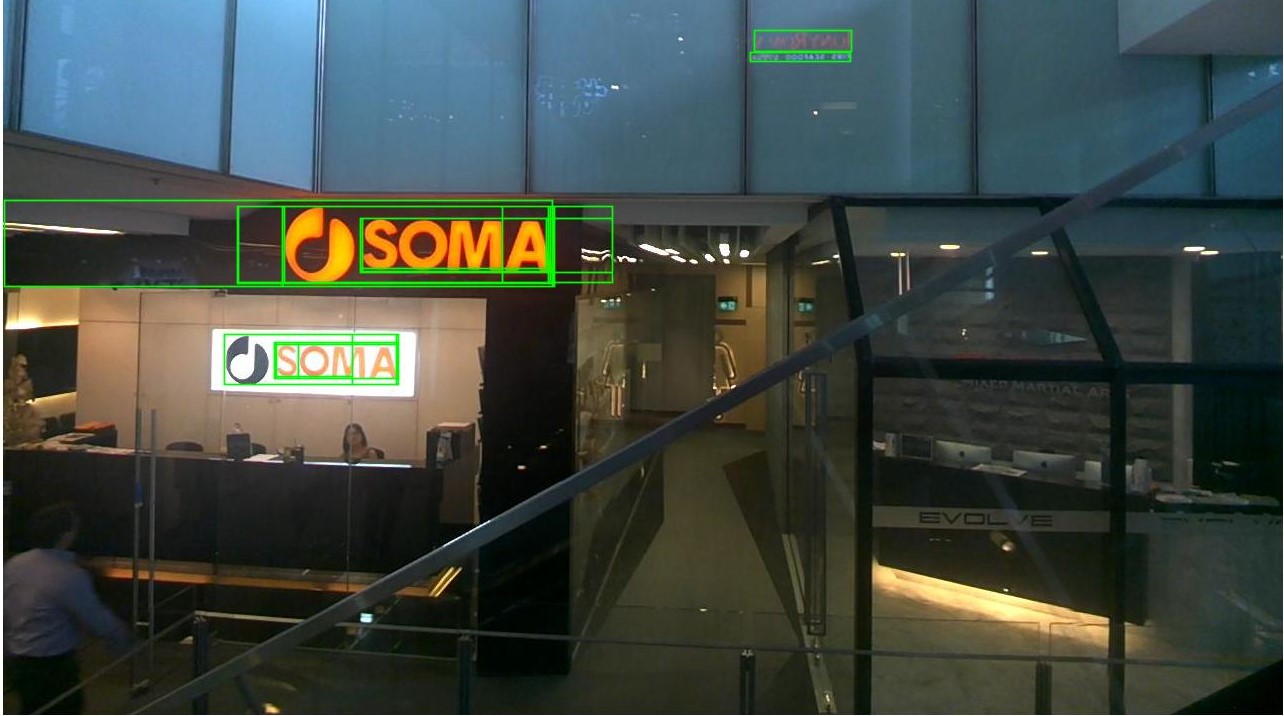} 
\end{tabular}
\end{center}
\caption{Top ranked regions from COCO-Text and ICDAR-Challenge4 using suppression strategy with a threshold of 0.10.}
\label{fig:suppression-COCO-icdar}
 	\end{minipage}\qquad
 	\begin{minipage}[t!]{.4\textwidth} 	
 		\centering
\includegraphics[height=2.80cm]{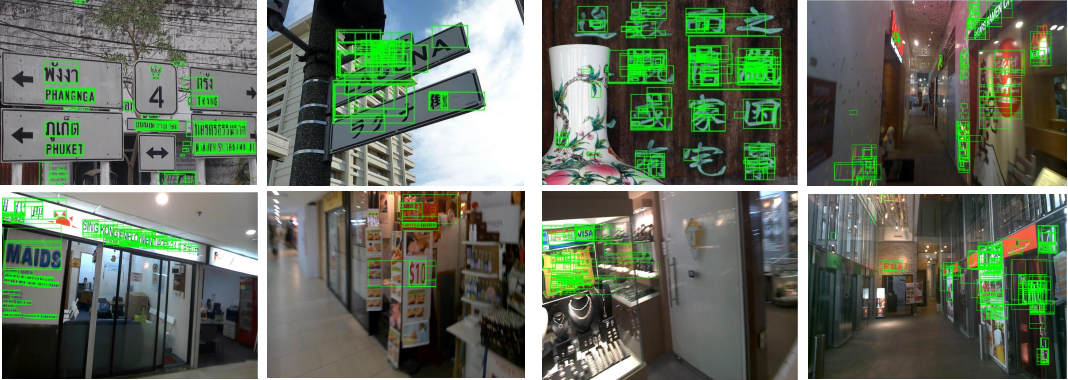}
\caption{Multi-language examples. Top ranked regions from COCO-Text and ICDAR-Challenge4 using suppression strategy with a threshold of 0.10 are shown.}
\label{fig:multi-language}
 	\end{minipage}
\end{figure*} 


\section{Conclusions}
\label{Conclusion}

In this paper we proposed an approach to improving the quality of a text-specific object proposals algorithm by exploiting local predictions from a Fully Convolutional Network trained to segment text from non-text regions. Our method re-scores text region proposals by integrating the pixel-wise text probability provided by the FCN model. This allows us to achieve high text region recall while considering significantly fewer candidate regions. Hence, it is capable of accelerating text detection. Experimental results show that our proposed approach achieves a performance superior to the baseline algorithm on COCO-Text and ICDAR2015-Challenge4 benchmarks. 
In the future, we plan to 
extend the proposed method to perform as an end-to-end text recognition system.


\section*{Acknowledgment}
This work was supported in part by the project TIN2014-52072-P and by the Eurecat Catalan Technology Center.

\end{document}